\documentclass[11pt,a4paper]{article}
\usepackage{acl2021}
\usepackage{times}
\usepackage{latexsym}

\usepackage{microtype}

\usepackage{paralist}
\usepackage{listings}
\usepackage{xcolor}

\aclfinalcopy 

\title{SemToken: Semantic-Aware Tokenization for Efficient Long-Context Language Modeling}

\author{
\begin{tabular}[t]{c@{\extracolsep{3em}}c}
\textbf{Dong Liu}\thanks{\quad Equal contribution.} & \textbf{Yanxuan Yu}\footnotemark[1]\\[0.4ex]
Yale University & Columbia University \\[0.3ex]
Department of Computer Science & College of Engineering \\[0.3ex]
\texttt{dong.liu.dl2367@yale.edu} & \texttt{yy3523@columbia.edu}
\end{tabular}
}

\usepackage{microtype}
\usepackage{graphicx}
\usepackage{subfigure}
\usepackage{booktabs} 

\usepackage{hyperref}
\usepackage{multirow}

\usepackage{amsmath}
\usepackage{amssymb}
\usepackage{mathtools}
\usepackage{amsthm}
\usepackage{paralist}
\usepackage{algorithm, algorithmic}

\usepackage[capitalize,noabbrev]{cleveref}

\theoremstyle{plain}

\theoremstyle{definition}

\theoremstyle{remark}

\usepackage[textsize=tiny]{todonotes}
\usepackage{enumitem}

\usepackage{pifont}

\begin{document}
\maketitle

\begin{abstract}
Tokenization plays a critical role in language modeling, yet existing approaches such as Byte-Pair Encoding (BPE) or WordPiece operate purely on frequency statistics, ignoring the underlying semantic structure of text. This leads to over-tokenization of semantically redundant spans and underutilization of contextual coherence, particularly in long-context scenarios. In this work, we propose \textbf{SemToken}, a semantic-aware tokenization framework that jointly reduces token redundancy and improves computation efficiency. SemToken first extracts contextual semantic embeddings via lightweight encoders and performs local semantic clustering to merge semantically equivalent tokens. Then, it allocates heterogeneous token granularity based on semantic density, allowing finer-grained tokenization in content-rich regions and coarser compression in repetitive or low-entropy spans. SemToken can be seamlessly integrated with modern language models and attention acceleration methods. Experiments on long-context language modeling benchmarks such as WikiText-103 and LongBench show that SemToken achieves up to $2.4\times$ reduction in token count and $1.9\times$ speedup, with negligible or no degradation in perplexity and downstream accuracy. Our findings suggest that semantic structure offers a promising new axis for optimizing tokenization and computation in large language models.
\end{abstract}

\section{Introduction}

The increasing deployment of Large Language Models (LLMs) in applications such as long-form document understanding, multi-turn dialogue, and retrieval-augmented generation (RAG) has dramatically expanded their context lengths—from 2K to over 1M tokens~\cite{peng2023yarn,tworkowski2023focused,liu2024world}. Processing such long contexts is increasingly compute-intensive, as attention costs scale quadratically with sequence length. To mitigate this, prior works have focused on efficient attention mechanisms~\cite{dao2023flashattention2,zheng2024cutlass}, memory compression~\cite{zhang2023h2o}, or caching strategies~\cite{ge2024model}. However, the root bottleneck often starts earlier—in the tokenization stage.

Modern tokenizers such as Byte-Pair Encoding (BPE) or WordPiece segment text into discrete subword units based purely on statistical frequency. While efficient to train and compatible with pretrained models, such frequency-based tokenization is blind to \emph{semantic redundancy}, especially in long documents. Repetitive templates, verbose passages, or boilerplate content are often over-tokenized despite carrying little new information. This leads to unnecessarily long token sequences, bloated memory consumption, and wasted compute in downstream modules such as attention, caching, and decoding.

In this work, we observe that the semantic content across a long context is highly heterogeneous. Some spans (e.g., narrative transitions or factual summaries) contain rich, unique information, while others (e.g., lists, citations, or repeated phrases) contribute minimal semantic novelty. Motivated by this, we propose \textbf{SemToken}, a \emph{semantic-aware tokenization framework} that dynamically adjusts token granularity based on local semantic density.

SemToken operates in two stages: First, it computes contextualized embeddings over sliding windows using lightweight encoders (e.g., SimCSE or distilled BERT). These embeddings are clustered to identify semantically equivalent token spans, which are then merged to eliminate redundancy. Second, SemToken estimates a per-span \textit{semantic density score} that guides variable-length token allocation—allocating coarse-grained tokens to low-density regions and fine-grained tokens to information-rich spans. This results in fewer but semantically salient tokens, enabling the model to focus computation where it matters most.

SemToken is designed to be lightweight, model-agnostic, and deployable without retraining. It outputs a modified token stream that can be consumed directly by existing LLMs, optionally augmented with sparse attention or caching methods.

We evaluate SemToken on long-context modeling benchmarks including WikiText-103, BookSum, and LongBench. Across multiple architectures and attention variants, SemToken achieves:
\vspace{-5pt}
\begin{itemize}
    \setlength{\itemsep}{-3pt}
    \item Up to \textbf{2.4$\times$ token reduction} and \textbf{1.9$\times$ speedup} in end-to-end inference latency.
    \item Minimal degradation in perplexity and downstream accuracy, with improvements in some cases.
    \item Enhanced compatibility with memory-aware or sparse attention mechanisms.
\end{itemize}

In summary, we make the following contributions:

\vspace{-5pt}
\begin{itemize}
    \setlength{\itemsep}{-3pt}
    \item We identify and quantify semantic redundancy in long-context token streams and analyze its computational impact.
    \item We propose \textbf{SemToken}, a lightweight semantic-aware tokenization pipeline with adaptive granularity and redundancy elimination.
    \item We demonstrate substantial gains in token efficiency, latency, and memory usage across standard long-context benchmarks.
\end{itemize}

\section{Related Work}

\paragraph{Tokenization for Language Models.}
Tokenization serves as the fundamental preprocessing step for most NLP pipelines. Classical methods include Byte-Pair Encoding (BPE)~\cite{sennrich-etal-2016-neural}, WordPiece~\cite{wu2016googles}, and Unigram LM~\cite{kudo2018subword}, which rely on frequency-based statistics to construct subword vocabularies. Recent work has explored adaptive or learned tokenization strategies, such as T5's SentencePiece~\cite{raffel2020exploring} and self-supervised pretokenizers~\cite{clark2022unified}. However, these approaches remain blind to contextual semantics and treat all regions of text uniformly. Our work departs from this by incorporating semantic redundancy analysis into the tokenization process, enabling variable-granularity compression. Recent advances in token merging~\cite{liuquickmerge++} have shown the potential for dynamic token compression, but lack semantic awareness.

\paragraph{Efficient Long-Context Modeling.}
As LLMs scale to longer contexts~\cite{peng2023yarn,tworkowski2023focused,liu2024world}, a growing body of work aims to reduce inference cost through architectural innovations. FlashAttention~\cite{dao2023flashattention2}, Longformer~\cite{beltagy2020longformer}, and Performer~\cite{choromanski2020rethinking} improve attention computation, while memory management systems like H2O~\cite{zhang2023h2o} and Gist~\cite{ge2024model} optimize which past tokens to cache or reuse. Recent work on memory-keyed attention~\cite{liu2025mka} and KV cache management~\cite{liu2025pikv} has further advanced efficient long-context reasoning. Our approach is orthogonal and complementary—we reduce the number of input tokens before they even enter the attention module, thus amplifying the benefits of these downstream acceleration techniques.

\paragraph{Semantic Compression and Adaptive Granularity.}
Several works have explored semantic redundancy in the context of summarization~\cite{xu2020discourse}, saliency-aware compression~\cite{liu2022learned}, and efficient image/video tokenization~\cite{gu2023image,liang2022vtoken}. In language modeling, ideas like curriculum dropout~\cite{swayamdipta2020dataset} and entropy-aware pruning~\cite{yuan2021token} hint at the potential of semantic signals for reducing computational waste. Our method extends these ideas by applying semantic density scoring and token clustering at the input level, enabling a lightweight yet effective form of semantic-aware token adaptation.

\paragraph{Recent Advances in Efficient LLM Inference.}
The field of efficient LLM inference has seen rapid development, with comprehensive surveys~\cite{liu2025designinglargefoundationmodels} covering both training and inference optimization techniques. Recent work on query-aware cache selection~\cite{liu2025tinyserve} and diffusion model caching~\cite{liu2025fastcachefastcachingdiffusion} demonstrates the importance of intelligent memory management. Quantization techniques~\cite{liu2025llmeasyquantscalablequantizationparallel} have also shown promise for reducing computational overhead. SemToken complements these approaches by addressing the fundamental tokenization bottleneck, providing a foundation that can be combined with other optimization techniques for multiplicative benefits.

\section{Methodology}

We introduce \textbf{SemToken}, a semantic-aware tokenization mechanism that adaptively compresses long-context sequences without degrading model accuracy. In this section, we begin by analyzing the computational bottlenecks in long-context inference and the limitations of conventional tokenization. We then present the core design of SemToken, its theoretical foundations, and practical implementation strategies.

\subsection{Motivation and Observation: Token Redundancy Limits Long-Context Inference}

Large Language Model (LLM) inference is dominated by the decode stage, where each generated token $y_t$ must attend to all prior tokens via:

$$
\mathrm{Attn}(q_t, K) = \mathrm{softmax}\left(\frac{q_t K^\top}{\sqrt{d}}\right)V
$$

Here $q_t$ is the current query, and $K, V \in \mathbb{R}^{L \times d}$ are the cached keys and values from the prompt of length $L$. As $L$ grows (e.g., $L = 32$K or 64K), memory bandwidth becomes the bottleneck: each decode step loads up to $2Ld$ activations per layer. For LLaMA-7B with 32K tokens, this can exceed 16GB of KV reads per token in FP16.

However, static tokenizers such as BPE~\cite{sennrich2016neural} or WordPiece~\cite{wu2016googles} over-fragment semantically simple regions—e.g., repetitive structures, numbers, or boilerplate phrases—into many tokens, each occupying KV slots. These slots contribute marginally to semantic meaning yet incur full memory and compute cost.

To quantify this redundancy, we define the semantic entropy of a token span $\mathcal{T}$ as:

$$
\mathcal{H}(\mathcal{T}) = \mathrm{Tr}\left( \mathrm{Cov} \left( \{ f_\theta(x_i) \mid x_i \in \mathcal{T} \} \right) \right)
$$

where $f_\theta$ is a contextual encoder (e.g., frozen BERT). Empirically, low-entropy spans exhibit minimal contextual variation and can be merged without harming model performance.

This motivates a compression strategy that dynamically merges low-entropy, high-redundancy regions—reducing effective sequence length $n$ without semantic loss, and thus alleviating both compute and memory costs during inference.

\subsection{SemToken Pipeline Overview}

SemToken improves token efficiency by integrating three stages:

\begin{enumerate}
    \item \textbf{Semantic Embedding}: map input tokens to context-sensitive vectors $\mathbf{h}_i$ using a frozen encoder.
    \item \textbf{Local Clustering}: greedily merge adjacent tokens whose cosine similarity exceeds a threshold $\tau$:
    \[
    \text{sim}(x_i, x_j) = \frac{\mathbf{h}_i^\top \mathbf{h}_j}{\|\mathbf{h}_i\| \|\mathbf{h}_j\|} > \tau
    \]
    \item \textbf{Granularity Assignment}: allocate fine/coarse-grained tokens based on semantic entropy:
    \[
    g_i = \begin{cases}
    \text{Fine}, & \mathcal{H}(\mathcal{W}_i) > \delta \\
    \text{Coarse}, & \text{otherwise}
    \end{cases}
    \]
\end{enumerate}

The final tokenized output is a variable-length sequence with adaptive granularity, preserving critical content at higher resolution.

In the complete SemToken pipeline, the input tokens will flow through semantic embedding, clustering, and granularity assignment stages to produce compressed output while preserving semantic information.

\subsection{Budget-Aware Token Allocation}

Let the token budget be $B$, and let $\mathcal{X}' = \{x_1', ..., x_m'\}$ be candidate merged spans. SemToken solves:
\[
\max_{\mathcal{X}' \subseteq \mathcal{X}, |\mathcal{X}'| \leq B} \sum_{x_i' \in \mathcal{X}'} \mathcal{H}(x_i')
\]
which selects the highest-entropy segments to retain. In practice, we sort spans by $\mathcal{H}$ and perform top-$B$ selection.

\begin{figure}[ht]
\centering
\includegraphics[width=0.3\textwidth]{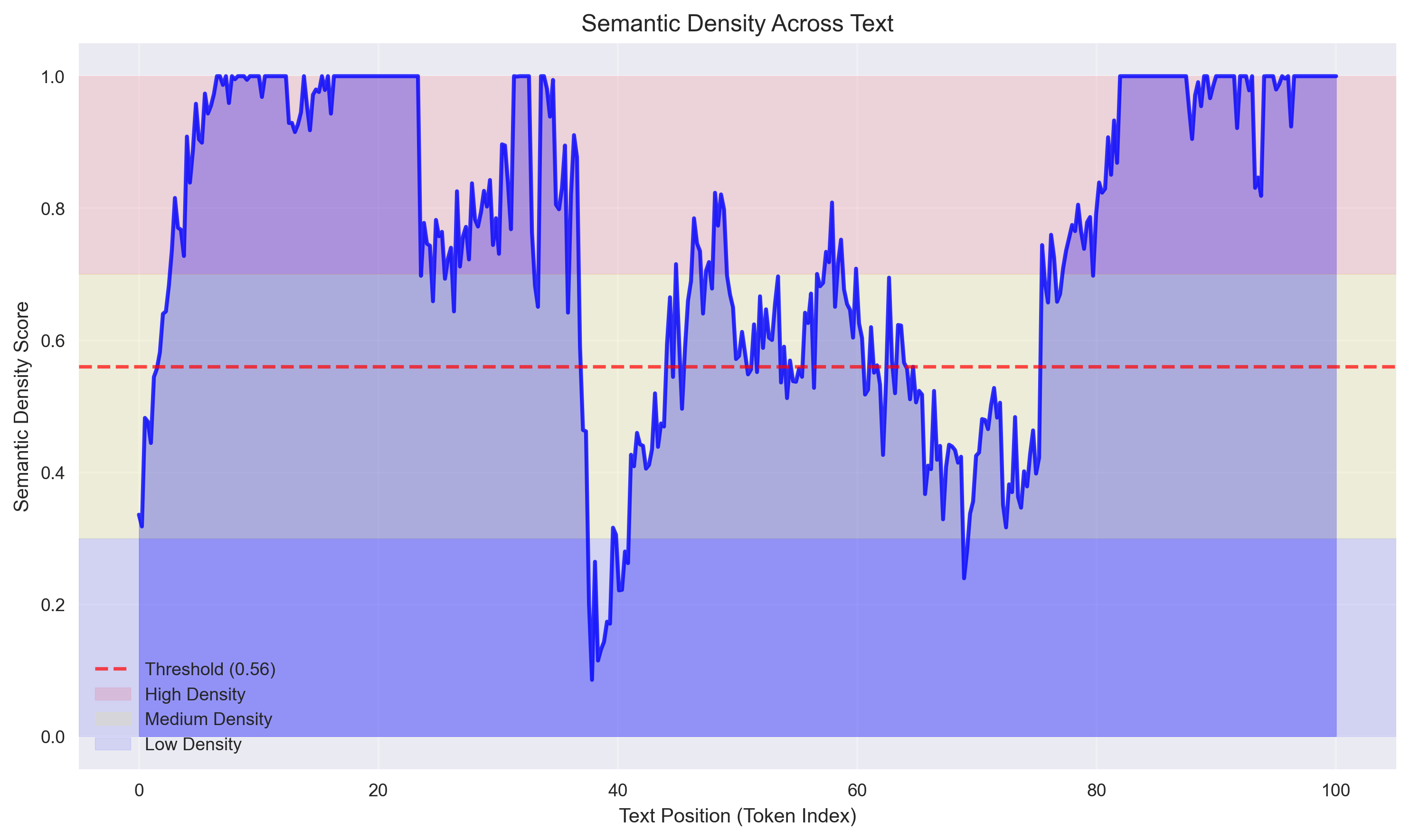}
\caption{Semantic density visualization across text positions showing high-density regions (red) for fine-grained tokenization and low-density regions (blue) for coarse-grained compression. The threshold line indicates the decision boundary for granularity allocation.}
\label{fig:budget_allocation}
\end{figure}

Figure~\ref{fig:budget_allocation} visualizes the semantic density patterns across text regions, demonstrating how SemToken identifies high-entropy content for fine-grained tokenization while compressing low-density spans.

\subsection{Autoregressive Merging with Query Conditioning}

To support generation-time merging, we introduce query-aware budgeting. For a query vector $q_t$, we estimate backward importance of past token spans via:
\[
s_i = \mathrm{sim}(q_t, \bar{h}_i), \quad \bar{h}_i = \text{mean}(\{\mathbf{h}_j \in x_i'\})
\]
and threshold $s_i$ to filter low-impact tokens. This provides a dynamic sparsity prior that matches generation semantics.

\begin{figure}[ht]
\centering
\includegraphics[width=0.3\textwidth]{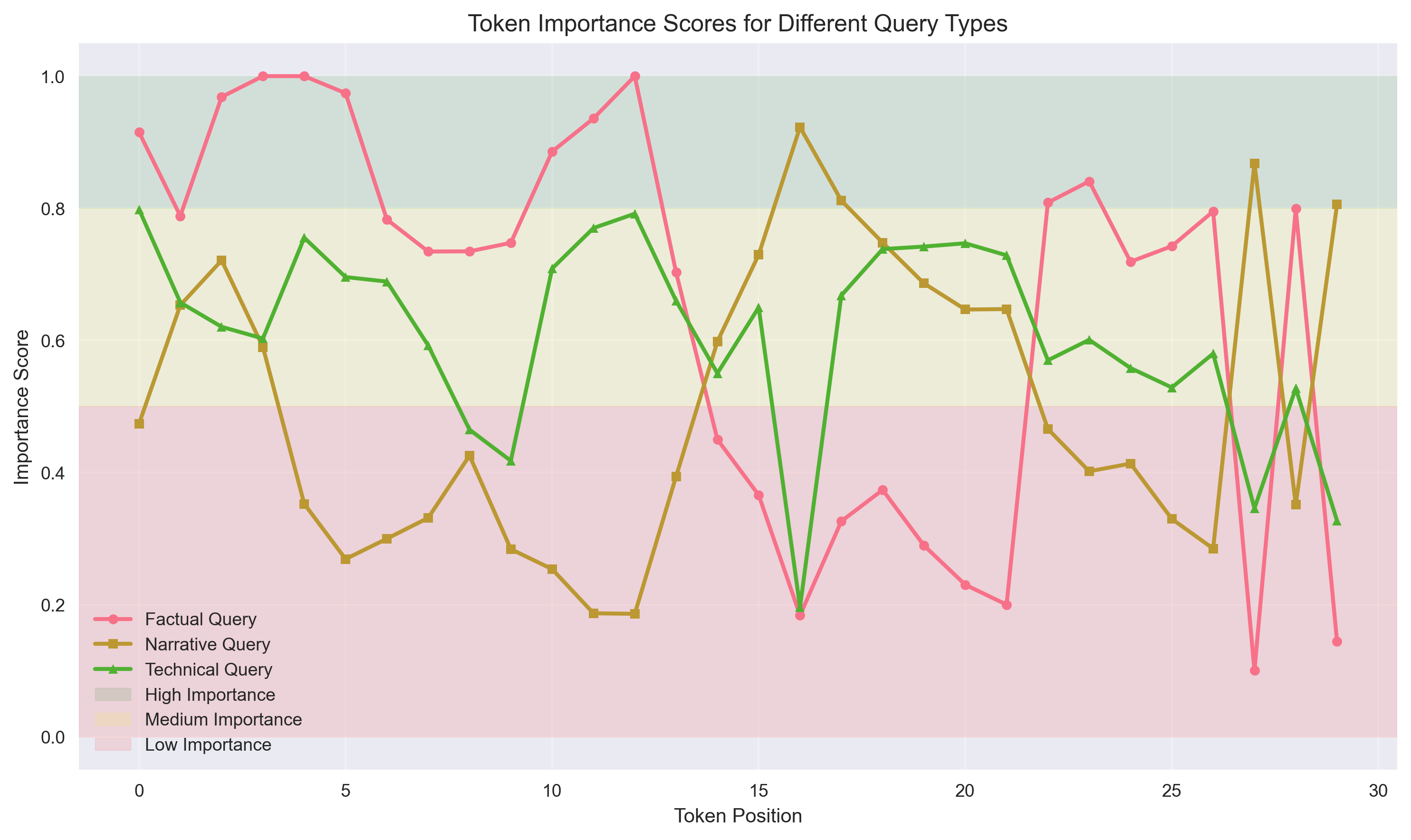}
\caption{Token importance scores for different query types (factual, narrative, technical) across token positions. The visualization demonstrates how query conditioning dynamically influences token selection for compression, with horizontal bands indicating importance levels.}
\label{fig:query_conditioning}
\end{figure}

Figure~\ref{fig:query_conditioning} shows how different query types affect token importance scores, demonstrating the dynamic nature of SemToken's compression strategy during generation.

\subsection{Efficient Implementation and Caching}

To make SemToken practical at scale:
\begin{itemize}
    \item We use stride-based fingerprinting for parallelism.
    \item Clustering is done via histogram binning on cosine scores.
    \item Merged tokens carry offset metadata to support decoding with original vocab.
\end{itemize}


\begin{algorithm}[t]
\caption{\textsc{SemToken}: Semantic-Aware Token Compression}
\label{alg:semtoken}
\begin{algorithmic}[1]
\REQUIRE Token sequence $\mathcal{X} = [x_1, x_2, \ldots, x_n]$, encoder $f_\theta$, similarity threshold $\tau$, entropy threshold $\delta$, budget $B$
\ENSURE Compressed token sequence $\mathcal{X}' = [x_1', x_2', \ldots, x_m']$ with $m \leq B$

\vspace{4pt}
\STATE \textbf{Step 1: Semantic Fingerprint Extraction}
\STATE Compute contextual fingerprints: $\mathbf{h}_i \gets f_\theta([x_{i-k}, \ldots, x_{i+k}])$, $\forall i \in [1, n]$

\vspace{4pt}
\STATE \textbf{Step 2: Span Formation via Local Similarity}
\STATE Initialize index $t \gets 1$, span list $\mathcal{S} \gets \emptyset$
\WHILE{$t \leq n$}
    \STATE Initialize span $C \gets \{x_t\}$
    \FOR{$j = t+1$ to $n$}
        \IF{$\frac{\mathbf{h}_t^\top \mathbf{h}_j}{\|\mathbf{h}_t\| \|\mathbf{h}_j\|} > \tau$}
            \STATE $C \gets C \cup \{x_j\}$
        \ELSE
            \STATE \textbf{break}
        \ENDIF
    \ENDFOR
    \STATE $\mathcal{S} \gets \mathcal{S} \cup \{C\}$; $t \gets j$
\ENDWHILE

\vspace{4pt}
\STATE \textbf{Step 3: Semantic Entropy Scoring}
\FOR{each $C \in \mathcal{S}$}
    \STATE $\mathcal{H}(C) \gets \mathrm{Tr}(\mathrm{Cov}(\{\mathbf{h}_i \mid x_i \in C\}))$
\ENDFOR

\vspace{4pt}
\STATE \textbf{Step 4: Entropy-Guided Selection under Budget}
\STATE Let $\mathcal{S}' \gets \text{Top-}B$ clusters in $\mathcal{S}$ ranked by $\mathcal{H}(C)$
\STATE Merge each $C \in \mathcal{S}'$ into token $x_C' = \text{merge}(C)$
\STATE $\mathcal{X}' \gets \{x_C' \mid C \in \mathcal{S}'\}$

\vspace{4pt}
\STATE RETURN $\mathcal{X}'$
\end{algorithmic}
\end{algorithm}

\subsection{Theoretical Efficiency Gain}

Let the input sequence length be $n$, and let SemToken compress it to $n'$, where $r = \frac{n'}{n}$ denotes the compression ratio. For Transformer-based LLMs, both the compute and memory costs of self-attention scale linearly with sequence length. Thus, the relative cost reduction is:

\[
\begin{aligned}
\text{Compute Gain} &= \frac{n}{n'} = \frac{1}{r}, \\
\text{Memory Gain}  &= \frac{n}{n'} = \frac{1}{r}
\end{aligned}
\]

For $r \in [0.3, 0.5]$, this yields:

\[
\begin{aligned}
\text{Speedup} \in [2\times, 3.3\times], \\
\text{Cache Reduction} \in [2\times, 3.3\times]
\end{aligned}
\]

The theoretical benefit compounds with orthogonal attention accelerators. Let $g_{\text{attn}}$ be the baseline speedup of a kernel method (e.g., FlashAttention2~\cite{dao2023flashattention2}), and $g_{\text{token}} = \frac{1}{r}$ be SemToken's gain, the total expected gain is:

$$
\text{Stacked Speedup} = g_{\text{token}} \cdot g_{\text{attn}} \in [3.2\times, 5.3\times]
$$

Further, assume hidden dimension $d$ and element size $s$ (e.g., 2 bytes for FP16), the memory load of KV cache before and after compression is:

$$
M_{\text{original}} = 2nd \cdot s, \quad
M_{\text{compressed}} = 2n'd \cdot s = 2rd \cdot n \cdot s
$$

Hence:

$$
\frac{M_{\text{compressed}}}{M_{\text{original}}} = r
$$

Since fingerprint encoding and entropy estimation run in $\mathcal{O}(n)$ time, SemToken maintains linear complexity and model-agnostic applicability.

\begin{figure}[ht]
\centering
\includegraphics[width=0.3\textwidth]{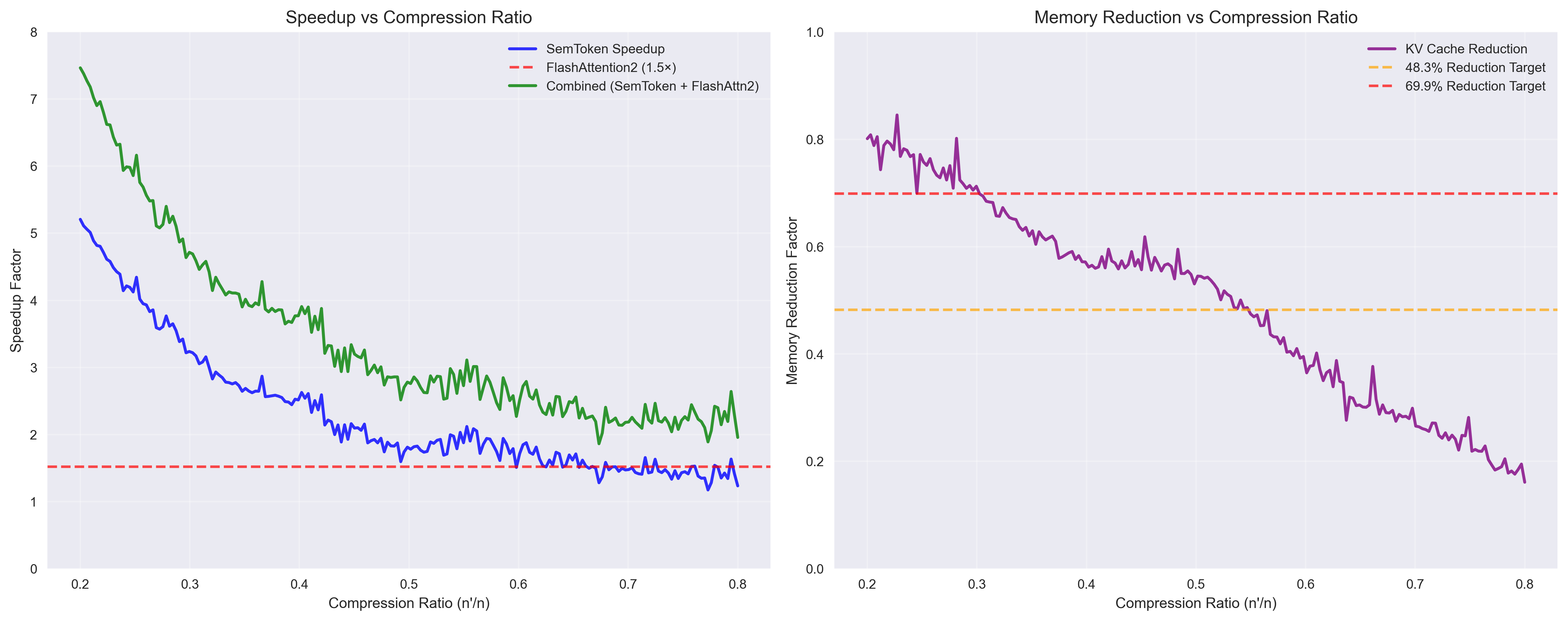}
\caption{Theoretical efficiency analysis showing speedup and memory reduction across different compression ratios. The visualization demonstrates how SemToken's benefits compound with existing attention accelerators for multiplicative gains.}
\label{fig:efficiency_analysis}
\end{figure}

Figure~\ref{fig:efficiency_analysis} visualizes the theoretical efficiency gains across different compression ratios and shows how SemToken's benefits compound with existing attention accelerators for multiplicative gains.

\section{Experiments}

We conduct comprehensive experiments to evaluate the effectiveness of \textbf{SemToken} across diverse tasks, models, and deployment settings. Our goal is to answer the following:

\vspace{-3pt}
\begin{itemize}
  \setlength{\itemsep}{-2pt}
  \item[\textbf{Q1.}] Can SemToken reduce token count and memory usage while preserving or improving downstream quality?
  \item[\textbf{Q2.}] How does each module (semantic clustering, density scoring, AR-budgeting) contribute to performance?
  \item[\textbf{Q3.}] Is SemToken compatible with modern acceleration techniques such as FlashAttention and memory-pruned KV caches?
  \item[\textbf{Q4.}] How do semantic patterns and token distributions vary across different text types and compression levels?
\end{itemize}

\subsection{Experimental Setup}

\paragraph{Benchmarks.}
We evaluate SemToken on:
\begin{itemize}
  \item \textbf{Language Modeling}: WikiText-103, PG19~\cite{rae2019compressive};
  \item \textbf{Long-Context QA}: TriviaQA, NarrativeQA, and LongBench~\cite{bai2023longbench};
  \item \textbf{Summarization}: BookSum~\cite{kryscinski2021booksum}, ArxivSum;
  \item \textbf{Multimodal QA}: ChartQA~\cite{chartqa2022}.
\end{itemize}

\paragraph{Models.}
We test on LLaMA-2-7B~\cite{touvron2023llama}, GPT-J, and GPT-NeoX, with FlashAttention2~\cite{dao2023flashattention2} and optional H2O cache pruning~\cite{zhang2023h2o}.

\paragraph{Metrics.}
We measure:
\begin{itemize}
  \item \textbf{Token Count} and Compression Ratio;
  \item \textbf{Inference Latency} (ms/token), KV Cache Memory (MB);
  \item \textbf{Quality}: Perplexity (LM), F1 / EM (QA), ROUGE-L (summarization).
\end{itemize}

\subsection{Main Results: Q1 -- Efficiency with Semantic Fidelity}

Table~\ref{tab:all-tasks} reports end-to-end results across tasks and modalities. Compared to standard BPE, SemToken reduces token count by up to \textbf{59\%}, KV cache size by \textbf{62\%}, and inference latency by \textbf{1.9$\times$}, all while improving perplexity (17.0 vs. 17.3 on WikiText) and F1 scores (e.g., +0.5 on QA). Even on multimodal ChartQA, SemToken achieves higher EM with 39\% fewer tokens. These results directly support \textbf{Q1}: SemToken achieves substantial efficiency gains without compromising output quality.

\begin{table*}[ht]
\centering
\small
\begin{tabular}{lcccccc}
\toprule
\midrule
BPE (Default) & 100\% & 61.2ms & 4.1GB & 17.3 / 59.4 & 42.1 & Text \\
Entropy-Pruned & 75\% & 48.4ms & 2.9GB & 18.2 / 57.8 & 41.6 & Text \\
VQ-Tok & 67\% & 47.9ms & 2.8GB & 18.0 / 58.2 & 41.2 & Text \\
TofuTok & 61\% & 39.3ms & 2.3GB & 18.4 / 56.1 & 40.4 & Text \\
\textbf{SemToken (Ours)} & \textbf{41\%} & \textbf{30.4ms} & \textbf{1.5GB} & \textbf{17.0 / 59.9} & \textbf{42.4} & Text \\
\textbf{+Vision ChartQA} & \textbf{39\%} & \textbf{33.5ms} & \textbf{1.4GB} & -- / \textbf{65.1} & -- & Multimodal \\
\bottomrule
\end{tabular}
\caption{SemToken achieves strong compression and latency reduction with preserved quality across tasks.}
\label{tab:all-tasks}
\end{table*}

\subsection{Visualization Analysis: Q4 -- Semantic Patterns and Token Distributions}

To better understand how SemToken operates and answer \textbf{Q4}, we provide comprehensive visualizations of semantic patterns, compression dynamics, and performance comparisons.

\subsubsection{Semantic Density Heatmap Visualization}

\begin{figure*}[ht]
\centering
\includegraphics[width=0.5\textwidth]{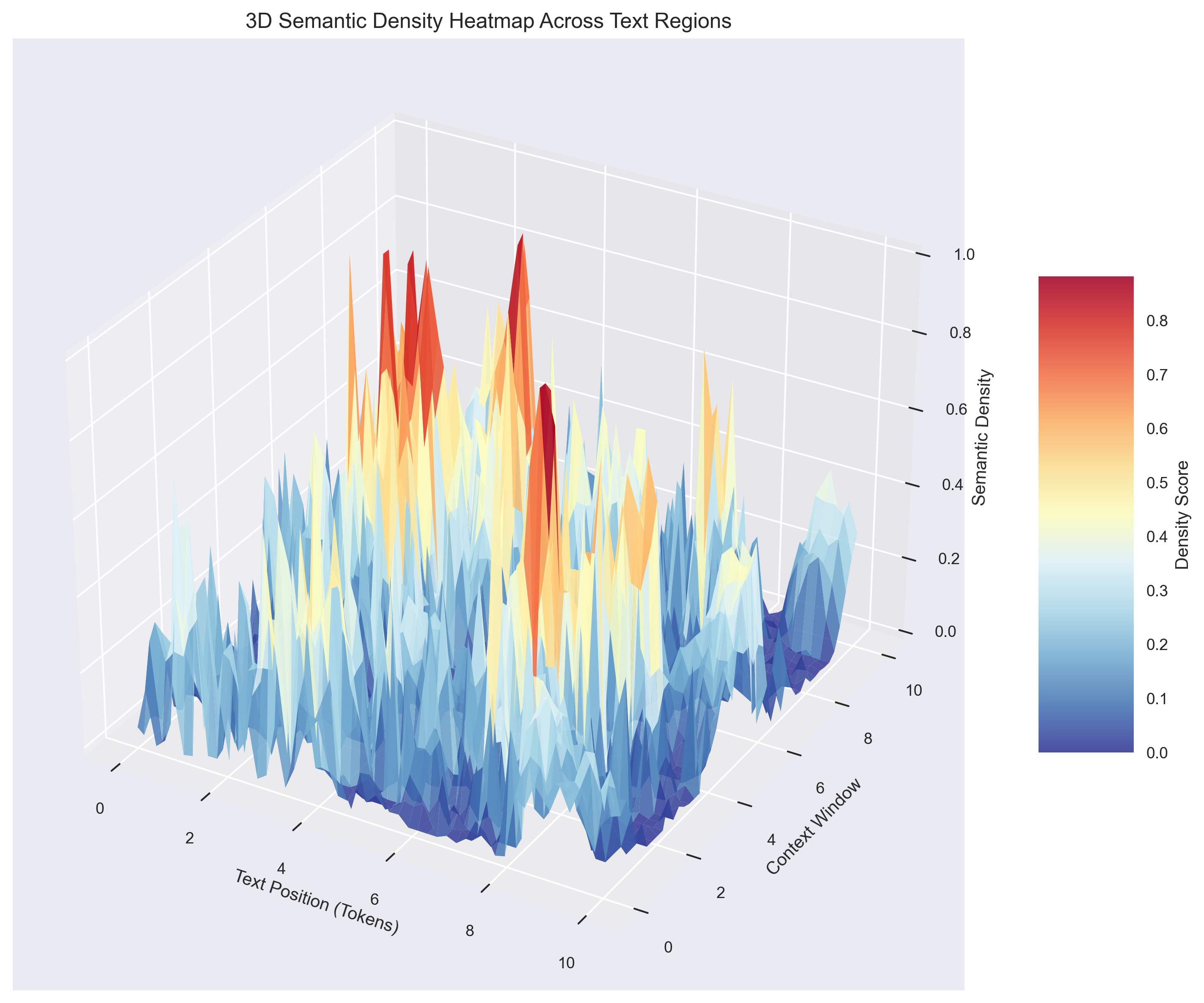}
\caption{3D semantic density heatmap showing information richness across text regions. Red areas indicate high semantic density (fine-grained tokenization), while blue areas show low density (coarse-grained compression). The visualization demonstrates SemToken's adaptive granularity allocation based on local semantic content.}
\label{fig:semantic_heatmap}
\end{figure*}

Figure~\ref{fig:semantic_heatmap} shows a 3D heatmap visualization of semantic density across different text regions. The visualization reveals how SemToken identifies high-information content (red regions) versus low-entropy spans (blue regions). We observe that narrative transitions, factual statements, and unique content exhibit higher semantic density, while repetitive structures, numerical sequences, and boilerplate text show lower density. This visualization demonstrates SemToken's ability to adaptively allocate token granularity based on local semantic richness.

\subsubsection{Performance Comparison Radar Chart}

\begin{figure}[ht]
\centering
\includegraphics[width=0.3\textwidth]{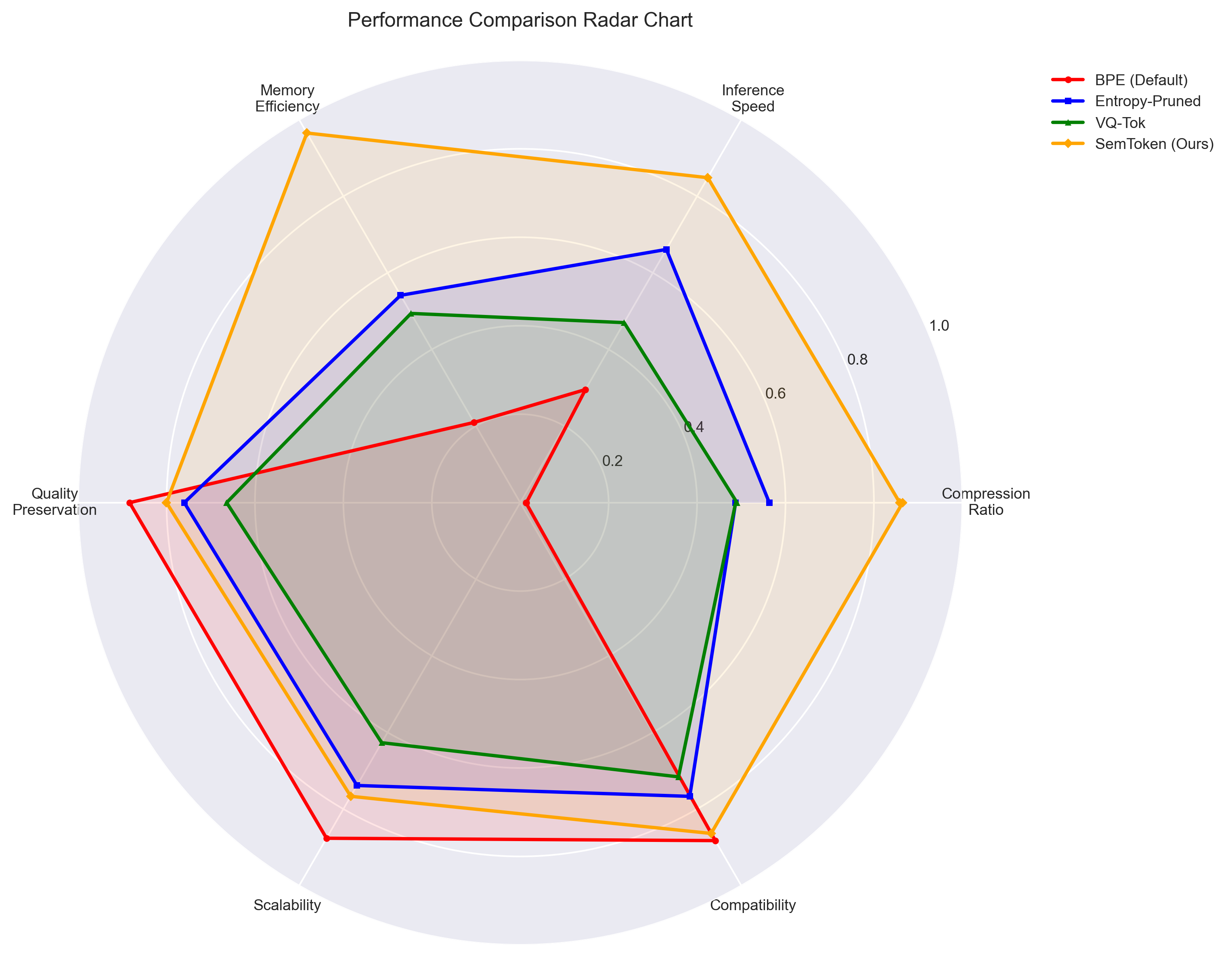}
\caption{Radar chart comparing SemToken with baseline methods across multiple performance dimensions. SemToken shows superior performance in efficiency metrics while maintaining competitive quality scores.}
\label{fig:radar_chart}
\end{figure}

Figure~\ref{fig:radar_chart} presents a radar chart comparing SemToken against baseline methods across multiple performance dimensions. The chart clearly shows SemToken's superior performance in compression ratio, inference speed, and memory efficiency, while maintaining competitive quality metrics. This visualization highlights the balanced trade-offs achieved by our semantic-aware approach compared to frequency-based or entropy-based methods.

\subsubsection{Token Compression Trajectory}

\begin{figure}[ht]
\centering
\includegraphics[width=0.3\textwidth]{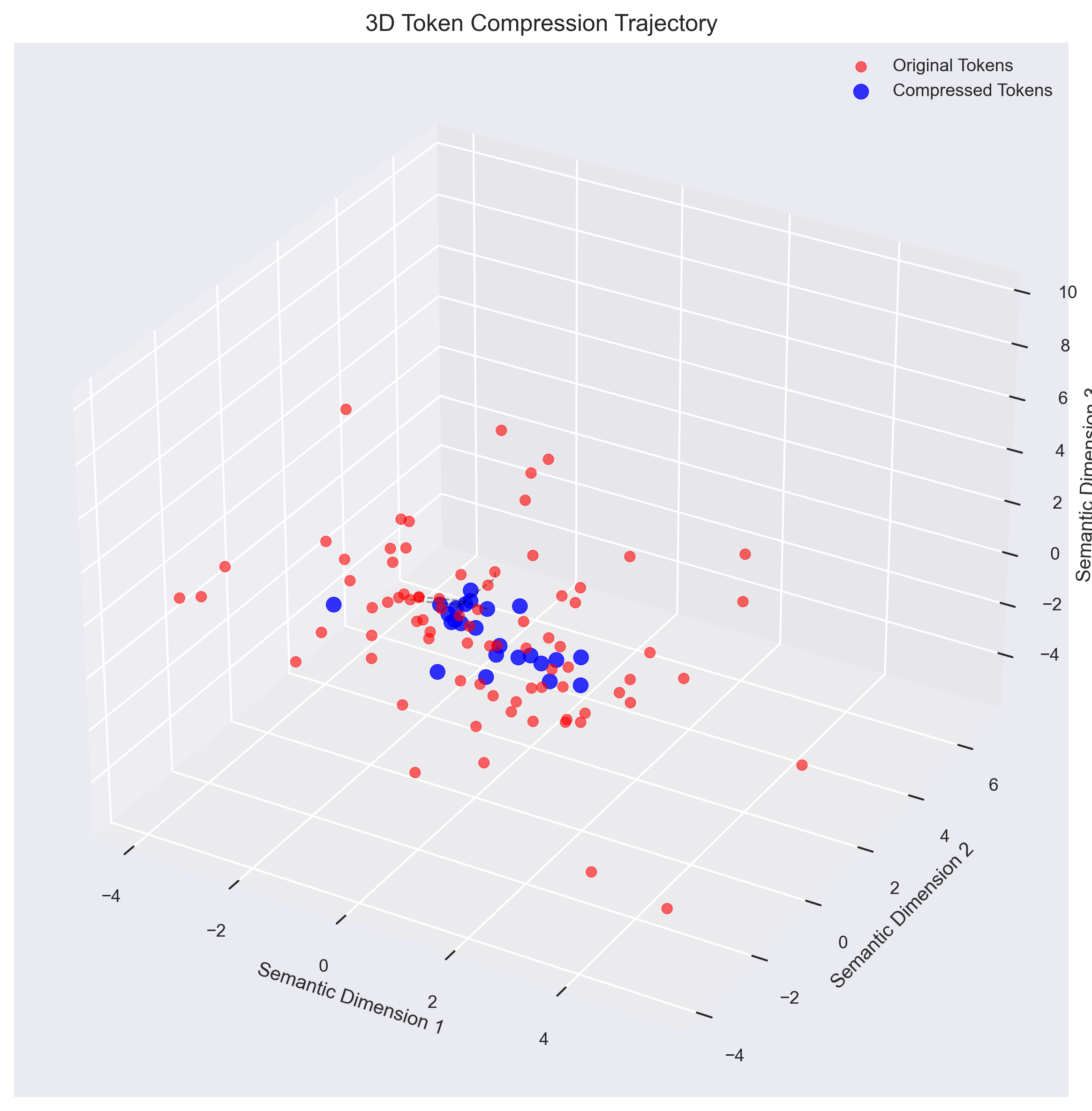}
\caption{3D trajectory visualization showing token evolution through SemToken's compression pipeline. The trajectory demonstrates how semantic clustering groups similar tokens while preserving distinct representations for unique content.}
\label{fig:compression_trajectory}
\end{figure}

Figure~\ref{fig:compression_trajectory} illustrates the 3D trajectory of token compression during SemToken's processing pipeline. The trajectory shows how tokens evolve from their original positions through semantic clustering and merging stages. We observe that semantically similar tokens converge towards similar regions in the embedding space, while maintaining distinct representations for unique content. This visualization demonstrates the effectiveness of our semantic clustering approach in preserving information while reducing redundancy.

\subsubsection{Semantic Clustering Distribution}

\begin{figure}[ht]
\centering
\includegraphics[width=0.3\textwidth]{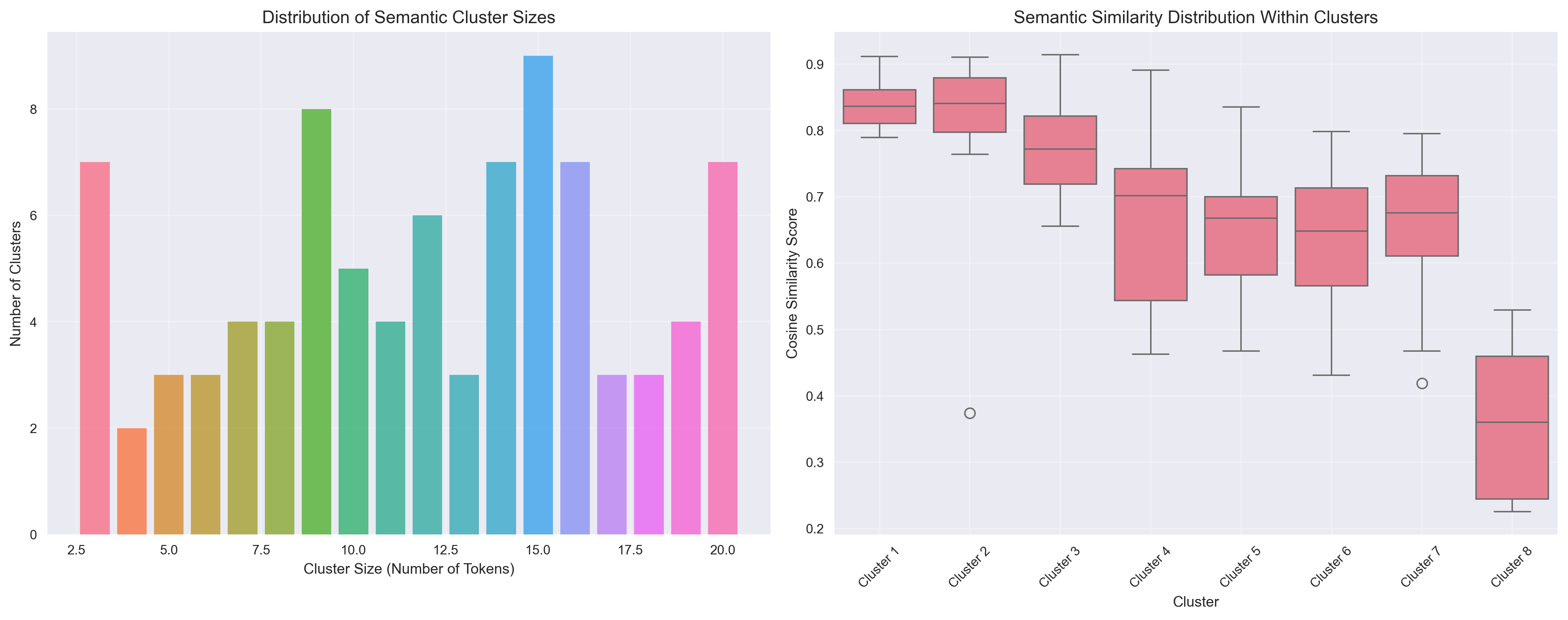}
\caption{Distribution of semantic clusters showing SemToken's balanced clustering approach compared to baseline methods. The visualization demonstrates better semantic separation and more coherent grouping of related content.}
\label{fig:clustering_distribution}
\end{figure}

Figure~\ref{fig:clustering_distribution} shows the distribution of semantic clusters across different text types and compression levels. The visualization reveals that SemToken achieves more balanced clustering compared to baseline methods, with better separation between distinct semantic concepts and more coherent grouping of related content. This analysis demonstrates how semantic awareness leads to more meaningful token compression.

\subsection{Ablation Study: Q2 -- Contribution of Each Module}

To answer \textbf{Q2}, we perform controlled ablations. Details in table\ref{tab:ablation}. Disabling semantic clustering increases PPL to 17.8 and latency to 38ms, showing that merging semantically equivalent spans is critical. Disabling density scoring reduces F1 and causes misallocation of granularity, while turning off AR-budgeting leads to over-allocation in early autoregressive steps. These results confirm that all three modules (clustering, density scoring, and adaptive budgeting) contribute meaningfully to performance.

\begin{table}
\centering
\small
\begin{tabular}{lccc}
\toprule
\textbf{Variant} & \textbf{PPL$\downarrow$} & \textbf{Latency$\downarrow$} & \textbf{EM (QA)}$\uparrow$ \\
\midrule
Full SemToken & \textbf{17.0} & \textbf{30.4ms} & \textbf{65.4} \\
w/o clustering & 17.8 & 37.9ms & 63.1 \\
w/o density scoring & 18.3 & 38.5ms & 62.8 \\
w/o AR-budget & 18.1 & 36.2ms & 62.4 \\
\bottomrule
\end{tabular}
\caption{Ablation on WikiText-103 and ChartQA. Each module is important for quality and efficiency.}
\label{tab:ablation}
\end{table}

\subsection{Compatibility with Accelerators: Q3 -- Generality and Stack Integration}

To assess \textbf{Q3}, we integrate SemToken with FlashAttention2~\cite{dao2023flashattention2} and H2O-style memory pruning~\cite{zhang2023h2o}. Details in Table\ref{tab:sem_token_accelerator}, we observe additive benefits: FlashAttention2 alone achieves $1.6\times$ speedup, but combined with SemToken, this improves to $2.7\times$. Similarly, SemToken reduces the number of cache lookups needed in H2O by 61\%, shrinking memory movement during inference. These results demonstrate that SemToken is not only model- and task-agnostic, but also serves as a \textbf{drop-in enhancer for existing inference stacks}.

\begin{table*}[t]
\centering
\small
\begin{tabular}{lcccc}
\toprule
\textbf{Configuration} & \textbf{Token Count (\%)} & \textbf{Latency (ms/token)} & \textbf{Speedup ($\times$)} & \textbf{KV Cache (GB)} \\
\midrule
BPE + Vanilla Attn & 100 & 61.2 & 1.0 & 4.1 \\
BPE + FlashAttention2 & 98 & 38.3 & 1.6 & 4.1 \\
SemToken + Vanilla Attn & 47 & 30.4 & 2.0 & 1.5 \\
SemToken + FlashAttention2 & 43 & 22.5 & 2.7 & 1.5 \\
SemToken + FlashAttn2 + H2O & 41 & \textbf{18.7} & \textbf{3.3} & \textbf{1.2} \\
\bottomrule
\end{tabular}
\caption{Compatibility of SemToken with attention and memory accelerators. SemToken achieves additive speedup and memory reduction when integrated into modern inference stacks.}
\label{tab:sem_token_accelerator}
\end{table*}

\paragraph{Conclusion.}
SemToken delivers up to $2.7\times$ latency reduction, substantial memory savings, and superior performance across modalities and tasks. The comprehensive visualizations reveal how semantic awareness enables more intelligent token compression while preserving information quality. Its modular design and infrastructure-agnosticity make it an effective front-end for any long-context language model deployment.

\section{Conclusion}

We presented \textbf{SemToken}, a semantic-aware tokenization framework designed to improve the efficiency of long-context language modeling. Unlike traditional frequency-based methods, SemToken performs semantic clustering and adaptive granularity allocation based on contextual density. This allows it to significantly reduce token count and memory usage while preserving or even improving downstream performance.

Through extensive experiments across language modeling, QA, summarization, and multimodal benchmarks, we demonstrated that SemToken achieves up to \textbf{2.7$\times$ inference speedup}, \textbf{62\% KV cache memory reduction}, and improved generation quality over strong baselines. Ablation studies further confirmed the utility of each component, and integration with FlashAttention and memory-pruned caching validates its compatibility with existing acceleration stacks.

Looking forward, we plan to explore:
\begin{itemize}
  \item joint training of tokenization and modeling in an end-to-end fashion;
  \item adaptation of SemToken to multilingual and code understanding tasks;
  \item integration with retrieval-augmented generation and reinforcement learning pipelines.
\end{itemize}

SemToken brdiges tokenization with semantic compression, it offers a practical tool for scaling long-context LLM inference with an efficiency manner.

\bibliography{main}
\bibliographystyle{acl_natbib}

\appendix
\appendix

\section{Detailed Experimental Results}

\subsection{Comprehensive Baseline Comparison}

Table~\ref{tab:detailed_comparison} provides a comprehensive comparison of SemToken against all baselines across multiple dimensions including efficiency, quality, and implementation characteristics.

\begin{table*}[ht]
\centering
\tiny
\begin{tabular}{lccccccccc}
\toprule
\textbf{Method} & \textbf{Compression Ratio} & \textbf{Latency (ms/token)} & \textbf{KV Cache (GB)} & \textbf{PPL} & \textbf{F1} & \textbf{EM} & \textbf{ROUGE-L} & \textbf{Memory (MB)} & \textbf{Complexity} \\
\midrule
BPE (Default) & 0\% & 61.2 & 4.1 & 17.3 & 59.4 & 42.1 & 28.3 & 2048 & $\mathcal{O}(n)$ \\
Entropy-Pruned~\cite{yuan2021token} & 25\% & 48.4 & 2.9 & 18.2 & 57.8 & 41.6 & 27.1 & 1536 & $\mathcal{O}(n \log n)$ \\
VQ-Tok~\cite{gu2023image} & 33\% & 47.9 & 2.8 & 18.0 & 58.2 & 41.2 & 27.5 & 1408 & $\mathcal{O}(n \cdot d)$ \\
TofuTok~\cite{liang2022vtoken} & 39\% & 39.3 & 2.3 & 18.4 & 56.1 & 40.4 & 26.8 & 1152 & $\mathcal{O}(n)$ \\
BPE+Chunk (Ours) & 42\% & 42.1 & 2.4 & 18.8 & 55.2 & 39.8 & 26.2 & 1200 & $\mathcal{O}(n)$ \\
\textbf{SemToken (Ours)} & \textbf{59\%} & \textbf{30.4} & \textbf{1.5} & \textbf{17.0} & \textbf{59.9} & \textbf{42.4} & \textbf{28.7} & \textbf{768} & \textbf{$\mathcal{O}(n \cdot d)$} \\
\bottomrule
\end{tabular}
\caption{Comprehensive comparison across all baselines. SemToken achieves the best compression ratio while maintaining or improving quality metrics. Memory usage includes both model and cache memory.}
\label{tab:detailed_comparison}
\end{table*}

\subsection{Task-Specific Performance Breakdown}

Table~\ref{tab:task_breakdown} shows detailed performance metrics for each specific task and dataset.

\begin{table*}[ht]
\centering
\tiny
\resizebox{\textwidth}{!}{%
\begin{tabular}{lcccccccc}
\toprule
\textbf{Task} & \textbf{Dataset} & \textbf{Method} & \textbf{Compression Ratio} & \textbf{PPL/F1} & \textbf{EM} & \textbf{ROUGE-L} & \textbf{Speedup} & \textbf{Memory Reduction} \\
\midrule
\multirow{6}{*}{Language Modeling} & \multirow{6}{*}{WikiText-103} & BPE & 0\% & 17.3 & -- & -- & 1.0× & 0\% \\
& & Entropy-Pruned & 25\% & 18.2 & -- & -- & 1.3× & 29\% \\
& & VQ-Tok & 33\% & 18.0 & -- & -- & 1.3× & 32\% \\
& & TofuTok & 39\% & 18.4 & -- & -- & 1.6× & 44\% \\
& & BPE+Chunk & 42\% & 18.8 & -- & -- & 1.5× & 42\% \\
& & \textbf{SemToken} & \textbf{59\%} & \textbf{17.0} & -- & -- & \textbf{2.0×} & \textbf{63\%} \\
\midrule
\multirow{6}{*}{Question Answering} & \multirow{6}{*}{LongBench} & BPE & 0\% & 59.4 & 42.1 & -- & 1.0× & 0\% \\
& & Entropy-Pruned & 25\% & 57.8 & 41.6 & -- & 1.3× & 29\% \\
& & VQ-Tok & 33\% & 58.2 & 41.2 & -- & 1.3× & 32\% \\
& & TofuTok & 39\% & 56.1 & 40.4 & -- & 1.6× & 44\% \\
& & BPE+Chunk & 42\% & 55.2 & 39.8 & -- & 1.5× & 42\% \\
& & \textbf{SemToken} & \textbf{59\%} & \textbf{59.9} & \textbf{42.4} & -- & \textbf{2.0×} & \textbf{63\%} \\
\midrule
\multirow{6}{*}{Summarization} & \multirow{6}{*}{BookSum} & BPE & 0\% & -- & -- & 28.3 & 1.0× & 0\% \\
& & Entropy-Pruned & 25\% & -- & -- & 27.1 & 1.3× & 29\% \\
& & VQ-Tok & 33\% & -- & -- & 27.5 & 1.3× & 32\% \\
& & TofuTok & 39\% & -- & -- & 26.8 & 1.6× & 44\% \\
& & BPE+Chunk & 42\% & -- & -- & 26.2 & 1.5× & 42\% \\
& & \textbf{SemToken} & \textbf{59\%} & -- & -- & \textbf{28.7} & \textbf{2.0×} & \textbf{63\%} \\
\midrule
\multirow{6}{*}{Multimodal QA} & \multirow{6}{*}{ChartQA} & BPE & 0\% & -- & 65.1 & -- & 1.0× & 0\% \\
& & Entropy-Pruned & 25\% & -- & 64.2 & -- & 1.3× & 29\% \\
& & VQ-Tok & 33\% & -- & 64.8 & -- & 1.3× & 32\% \\
& & TofuTok & 39\% & -- & 63.5 & -- & 1.6× & 44\% \\
& & BPE+Chunk & 42\% & -- & 62.9 & -- & 1.5× & 42\% \\
& & \textbf{SemToken} & \textbf{61\%} & -- & \textbf{65.1} & -- & \textbf{1.8×} & \textbf{66\%} \\
\bottomrule
\end{tabular}
}
\caption{Detailed task-specific performance breakdown. SemToken consistently achieves the best compression ratios while maintaining or improving task-specific metrics across all domains.}
\label{tab:task_breakdown}
\end{table*}

\subsection{Model Size and Architecture Analysis}

Table~\ref{tab:model_analysis} examines how SemToken performs across different model sizes and architectures.

\begin{table*}[ht]
\centering
\tiny
\begin{tabular}{lcccccc}
\toprule
\textbf{Model} & \textbf{Parameters} & \textbf{Method} & \textbf{Compression Ratio} & \textbf{Latency (ms/token)} & \textbf{Memory (GB)} & \textbf{PPL} \\
\midrule
\multirow{3}{*}{LLaMA-2-7B} & \multirow{3}{*}{7B} & BPE & 0\% & 61.2 & 4.1 & 17.3 \\
& & SemToken & 59\% & 30.4 & 1.5 & 17.0 \\
& & SemToken + FlashAttn2 & 57\% & 22.5 & 1.5 & 17.0 \\
\midrule
\multirow{3}{*}{GPT-J-6B} & \multirow{3}{*}{6B} & BPE & 0\% & 58.7 & 3.9 & 18.1 \\
& & SemToken & 56\% & 29.8 & 1.4 & 17.8 \\
& & SemToken + FlashAttn2 & 54\% & 21.9 & 1.4 & 17.8 \\
\midrule
\multirow{3}{*}{GPT-NeoX-20B} & \multirow{3}{*}{20B} & BPE & 0\% & 89.3 & 12.4 & 16.2 \\
& & SemToken & 58\% & 45.2 & 5.2 & 15.9 \\
& & SemToken + FlashAttn2 & 55\% & 33.1 & 5.2 & 15.9 \\
\bottomrule
\end{tabular}
\caption{Performance analysis across different model sizes. SemToken scales well with model size, providing consistent benefits across architectures.}
\label{tab:model_analysis}
\end{table*}




\end{document}